\documentclass[runningheads]{llncs}

\usepackage{graphicx}
\usepackage{xcolor}
\usepackage{url}
\usepackage{amsmath,amssymb}
\usepackage{mathtools}
\usepackage{bm}
\usepackage{xspace}
\usepackage{microtype}
\usepackage{booktabs}

\usepackage[pagebackref=true,breaklinks=true,colorlinks,bookmarks=false,citecolor=green!80!black,linkcolor=red!80!black,urlcolor=blue]{hyperref}
\usepackage{cleveref}
\usepackage{subcaption}

\usepackage{csquotes}
\usepackage{siunitx}

\crefname{section}{Sec.}{Sections}
\crefname{figure}{Fig.}{Figure}
\crefname{table}{Tab.}{Table}
\crefname{equation}{Equ.}{Equation}

\newcommand{\onedot}{.\xspace}

\newcommand{\cf}{cf\onedot}
\newcommand{\ie}{i.\,e.,\xspace}

\renewcommand{\O}[1]{\ensuremath{\mathcal{O}(#1)}}

\begin{document}

\title{Non-deterministic Behavior of Ranking-based Metrics when Evaluating Embeddings}
\titlerunning{Non-deterministic Behavior of Ranking-based Metrics}

\author{Anguelos Nicolaou\inst{1,2} \and
Sounak Dey\inst{1} \and \\
Vincent Christlein\inst{2} \and 
Andreas Maier\inst{2} \and 
Dimosthenis Karatzas\inst{1}}

\authorrunning{A.\ Nicolaou et al.}

\institute{Computer Vision Center, Edificio O, Campus UAB, 08193 Bellaterra, Spain\\ 
\email{\{anguelos,sdey,dimos\}@cvc.uab.cat}\\ \and
 Pattern Recognition Lab, Friedrich-Alexander-Universit\"at Erlangen-N\"urnberg
\email{\{anguelos.nikolaou,vincent.christlein,andreas.maier\}@fau.de}}

\maketitle

\begin{abstract}
Embedding data into vector spaces is a very popular strategy of pattern recognition methods.
When distances between embeddings are quantized, performance metrics become ambiguous.
In this paper, we present an analysis of the ambiguity quantized distances introduce and provide bounds on the effect.
We demonstrate that it can have a measurable effect in empirical data in state-of-the-art systems.
We also approach the phenomenon from a computer security perspective and demonstrate how someone being evaluated by a third party can exploit this ambiguity and greatly outperform a random predictor without even access to the input data.
We also suggest a simple solution making the performance metrics, which rely on ranking, totally deterministic and impervious to such exploits.

\keywords{Deterministic \and
mAP \and
Ranking \and
Performance Evaluation \and
Word spotting \and
Precision \and
Recall \and
Adversarial
}
\end{abstract}

\section{Introduction}
\subsection{Motivation}
In typical pattern recognition works, researchers introduce algorithms and demonstrate their performance 
using specific metrics. 

The measure of difference between the obtained outputs and the given ideal output (ground-truth) is the actual estimated performance.
Although the examined systems can have some randomness, it is assumed they follow a statistical distribution, whose mean can be estimated; it is generally also assumed that measuring the difference from the ground-truth with metrics defined in the literature is purely deterministic.
However, while experimenting with a typical retrieval method, it occurred that a specific output from a system would yield  different measured performances when evaluated with data in a different order.
After investigation, the problem was found to be a form of numerical instability, possibly attributed to the 32 bit limitation of modern GPU computing.
The problem is quite general and affects performance evaluation metrics that require sorting distances matrices.
In this paper, we (1) reproduce the incoherence in the recorded behavior of a real-world system, (2) we provide a data-driven analysis of the phenomenon, and (3) provide a simple fix that provably makes ranking-based metrics, such as mean average precision (mAP)~\cite{larson2010introduction}, behave deterministically under these conditions.

\subsection{Performance Evaluation and Competitions}
Evaluation protocols are a sensitive matter.
As there is no way to assess them objectively, their validity is mostly determined by consensus on how informative they are.
In order to constrain experimental bias, the development of a system and evaluating a system is considered as two distinct acts and researchers always try to keep the two roles as distinct as possible.
In the case of competitions, the two roles are strictly segregated between participants and organizers.
For such reasons, competitions set the gold standard in performance evaluation and have gained popularity.
As an indication of the rising popularity and importance of competitions, in the context of ICDAR in 2017, \num{25} different competitions were hosted by different groups
while ten years earlier in 2007, there were only 3.
Competitions establish a good practice in performance evaluation, which people then apply to measure their own methods' performance. 

\section{Rank Based Metrics}
Most of the popular performance metrics associated with Information Retrieval (IR) are closely related among each other.
In the most usual form, IR systems return data from a database sorted by relevance with respect to a query sample.
Then, evaluation metrics are computed that assess the ranking.

Any classification problem can be considered a retrieval problem where all samples in the database having the same class as the query are the relevant documents.
In this way, ranking metrics have become prevalent in evaluating classification tasks.
\subsection{Metric Estimation}\label{metricestimation}
The first step in computing ranking metrics of an embedding is to compute the distance matrix $D \in \mathbb{R}^{Q\times K}$ between any query $x_q, q \in \{1,\ldots,Q\}$ and a database of size $K$ for a specified distance metric, or equivalently a similarity matrix.
The next step is to compute a relevance matrix
$R \in \{0,1\}^{Q\times K}$ with elements:
\begin{equation}
R_{q,k}=\begin{cases}
			1 & \text{if } y(x_q) = y(x_d)\\
            0 & \text{if } y(x_q) \neq y(x_d)
		\end{cases}\;,
\label{eq:relevance}
\end{equation}
where $y(x)$ denotes the class of sample $x$. 

The row-wise sorting of $R$ by the values in $D$, results in the so called \emph{correct matrix} $C\in \mathbb{R}^{Q\times K}$, containing elements that are $1$ if the $k$-th closest sample of the database to the query $q$ is relevant and $0$ otherwise. 
The matrix $C$ can be directly used to compute the precision and recall for any query $x_q$ and rank $k$ resulting in the matrix $Pr$ and $Rc$, respectively with elements: 
\begin{equation}
Pr_{q,k}= \frac{1}{k}\sum_{n=1}^{k} C_{q,n}\qquad Rc_{q,k}= \frac{\sum_{n=1}^{k} C_{q,n}}{\sum_{n=1} C_{q,n}}\;.
\label{eq:precision}
\end{equation}
These two matrices can be used to produce all established metrics such as mean Average Precision (mAP), precision at rank 10, accuracy etc.
In the remaining of this paper we focus on mAP as it is by far the most popular of these metrics but the observations and analysis can easily be extended to all such metrics.

\subsection{Performance Evaluation of Embeddings}
A deployable retrieval system can be defined as a system that ranks a database of samples with respect to a query sample.
In most cases, retrieval systems consist of two steps: (1) mapping samples into a representation, \ie typically a vector of fixed dimensionality and (2) returning the distance of the two representations.
Embedding methods map any sample to a metric space $\mathbb{R}^n$ and usually used in combination with a metric distance, such as Euclidean distance, to form a retrieval system.

Performance evaluation protocols should be designed such
% I think this is a very special case and can be omitted
%that even if the same person developed the system and evaluates it, 
any person acts either as a \emph{creator} (of a method) or an \emph{evaluator} (of the method) at any given time.
It follows that the test-set should not be 
%acceptable 
accessible to the creator, %developer,
%and in case they are the same person, the creator should finalise his work before the evaluator starts his.
%%Non plus ultra
Ideally, the only thing the creator should pass to the evaluator is an opaque system (black-box) that produces outputs for given inputs and the evaluator should run this system on sequestered data and return a performance score.
The aforementioned is the higher standard for performance evaluation. Yet, quite often, the creator receives the test samples and reports the outputs instead of providing his system as a black-box; yet any evaluation protocol should be designed so that it can accommodate the strict separation of creator and evaluator.

The question arises, where does the retrieval system end and where does the evaluation system begin?
Under the assumption that the embedding method has a high cost, from an evaluators perspective, it is a lot faster to compute all the embeddings in the test-set only once, and then compute the distance matrix of them given a distance metric.
The other alternative, ranking the database for every query sample independently, would cost a lot more and is practically intractable, because for each query, the embeddings for the database are recomputed.

In terms of complexity, assuming the number of queries and the size of the retrieval database to be of approximately the sane size $m$, the dimensionality of the embedding to be a constant and the cost of mapping a single sample as $k$, then the complexity of a performance evaluation for systems with ranked outputs can be given by:
\begin{equation}
\label{eq:complexity1}
\O{k \times m \times m} = \O{km^2}
\end{equation}

On the other hand, under the assumption that the embedding dimensionality $n$ is a constant, the complexity for evaluating a black-box producing embeddings is given by:
\begin{equation}
\label{eq:complexity2}
\O{k \times m + k \times m + m \times m} = \O{max(m^2,km)}
\end{equation}

The aforementioned computability issue is not only theoretical, it is also well exemplified in the evolution the writer identification competition from ICHFR 2012~\cite{louloudis2012icfhr} to ICDAR 2013~\cite{louloudis2013icdar} where the increase of  the test set made the transition inevitable 
The above problem is not just about evaluating, retrieval methods, such as word-spotting with dynamic time-warping~\cite{rath2003word}, are not tractable for large-scale retrieval systems due to their computational complexity. Embedding methods ensure a tractable computational complexity for retrieval system and thus, they are so important.

\subsection{Equidistant Samples}
When sorting is part of the evaluation protocol, each sorting of the database that affects the ranking of relevant samples must be deterministic and unambiguous.
If rows in the distance matrix contain duplicate distances, then ranking becomes undefined.
Important to note is that an undefined algorithmic behavior in the context of sorting, is not the same as a random one. 
As opposed to random behavior, we cannot obtain an estimate of the expectation by repeatedly running the algorithm.
Worse than that, undefined behavior might behave deterministically with respect to unknown and theoretically irrelevant factors, such as the order of the queries or even memory availability.
This allows for bugs that are hard to detect and hard to be reproduced.

\begin{figure}[tb]
\includegraphics[width=\textwidth]{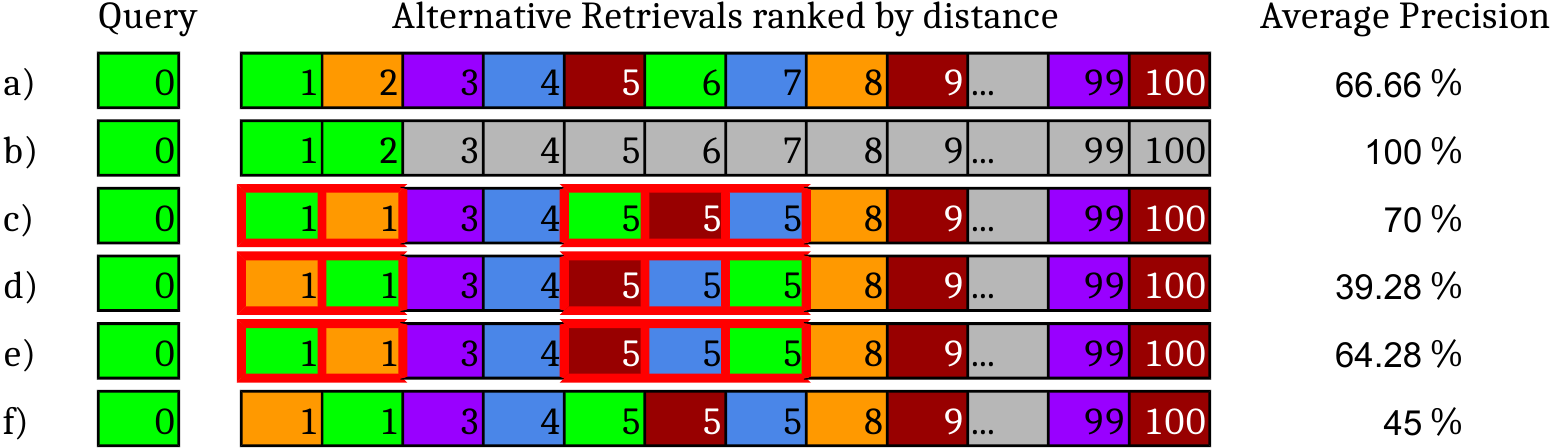}
\caption{mAP calculated for a query where there are 2 relevant samples out of \num{100} in the database. Green boxes are the relevant samples, non-green colored boxes are samples of other classes, while gray boxes signify irrelevant boxes whose class doesn't matter. Red borders denote clusters of samples that are equidistant from the query.} 
\label{fig:map_example}
\end{figure}

In \cref{fig:map_example} the problem and the effect equidistant samples in the database have on the resulting mAP is demonstrated.
Line a) is an indicative retrieval given a query where the relevant data appears in the first and sixth position.
Line b) represents a perfect retrieval case, which obtains an AP of \SI{100}{\percent}.
Lines c) to f) represent alternative sorting of the same embeddings where the first two samples and the fifth to seventh samples are equidistant from the query.
As can be seen, ambiguity only occurs when consecutive equidistant samples are both relevant (green) and non relevant (non-green).

Other metrics depending on sorting are also affected to an equivalent degree, but we focus on mAP because it is the most popular metric. 

\section{Experimental Data Analysis}\label{experiments}

\subsection{PHOCNET}
Although this paper addresses the general case of vector embeddings, we show experimental validation using a specific retrieval task known as segmented word-spotting.
Segmented word-spotting classifies a word-image into word classes, \ie elements of a dictionary.
Word-spotting is quite often modeled as a typical embedding system used in the context of information retrieval.
Domain adaptation has been a popular strategy, allowing to learn embeddings which map both word-images and word-transcriptions into a common subspace, the Pyramid Histogram Of Characters (PHOC)~\cite{almazan2014word}. 

\begin{figure}[tb]
\centering
\includegraphics[width=.32\textwidth]{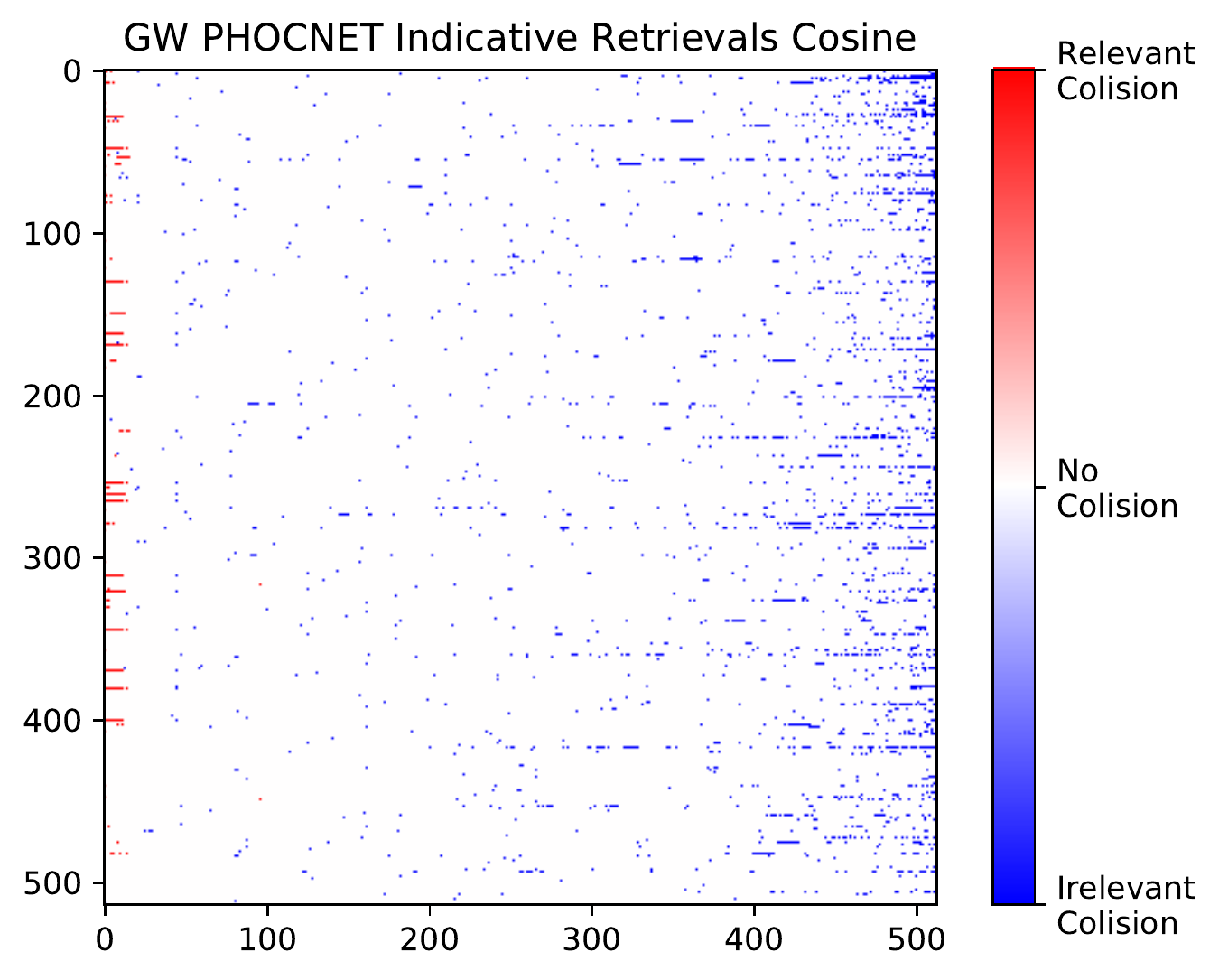}
\includegraphics[width=.32\textwidth]{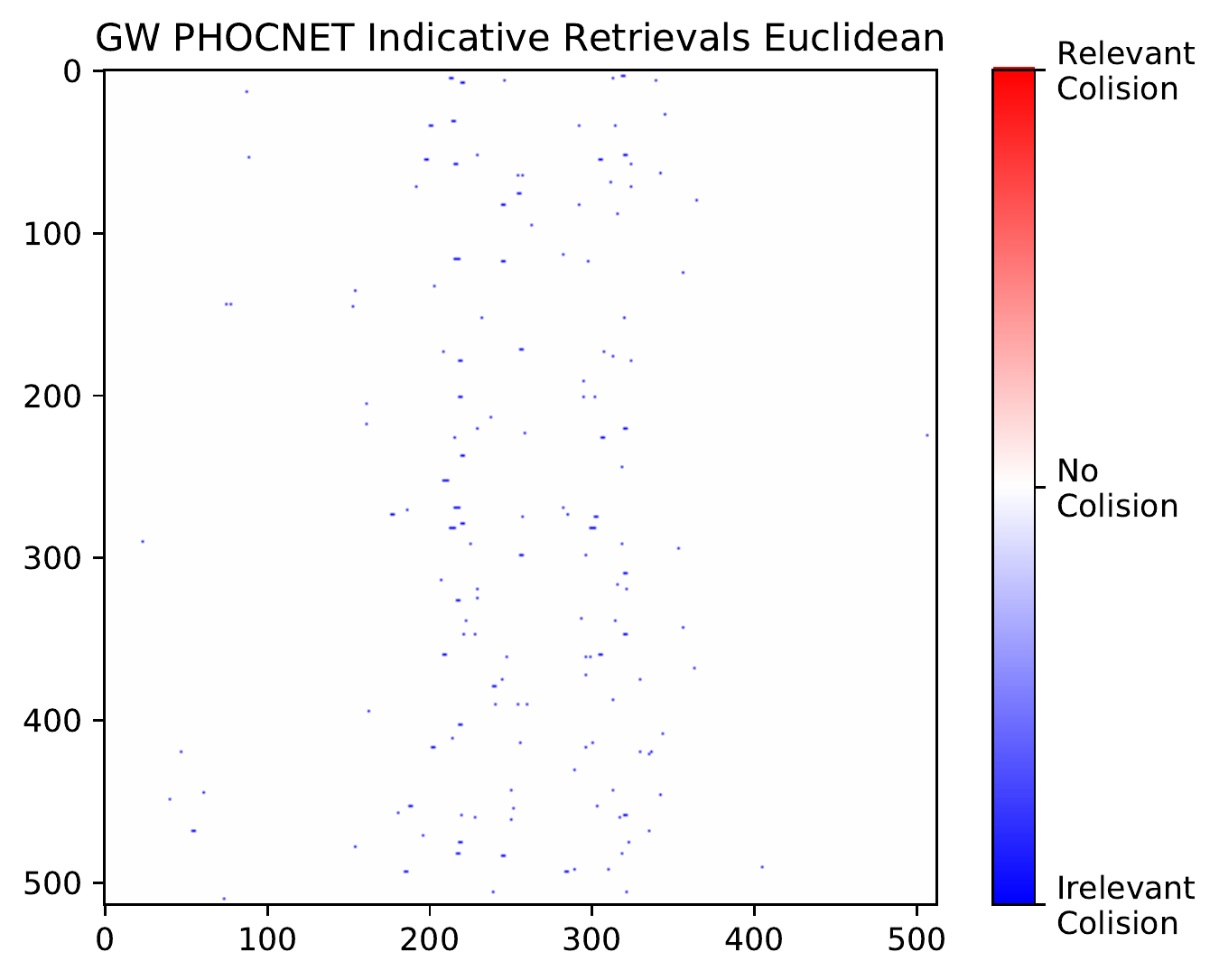}
\includegraphics[width=.32\textwidth]{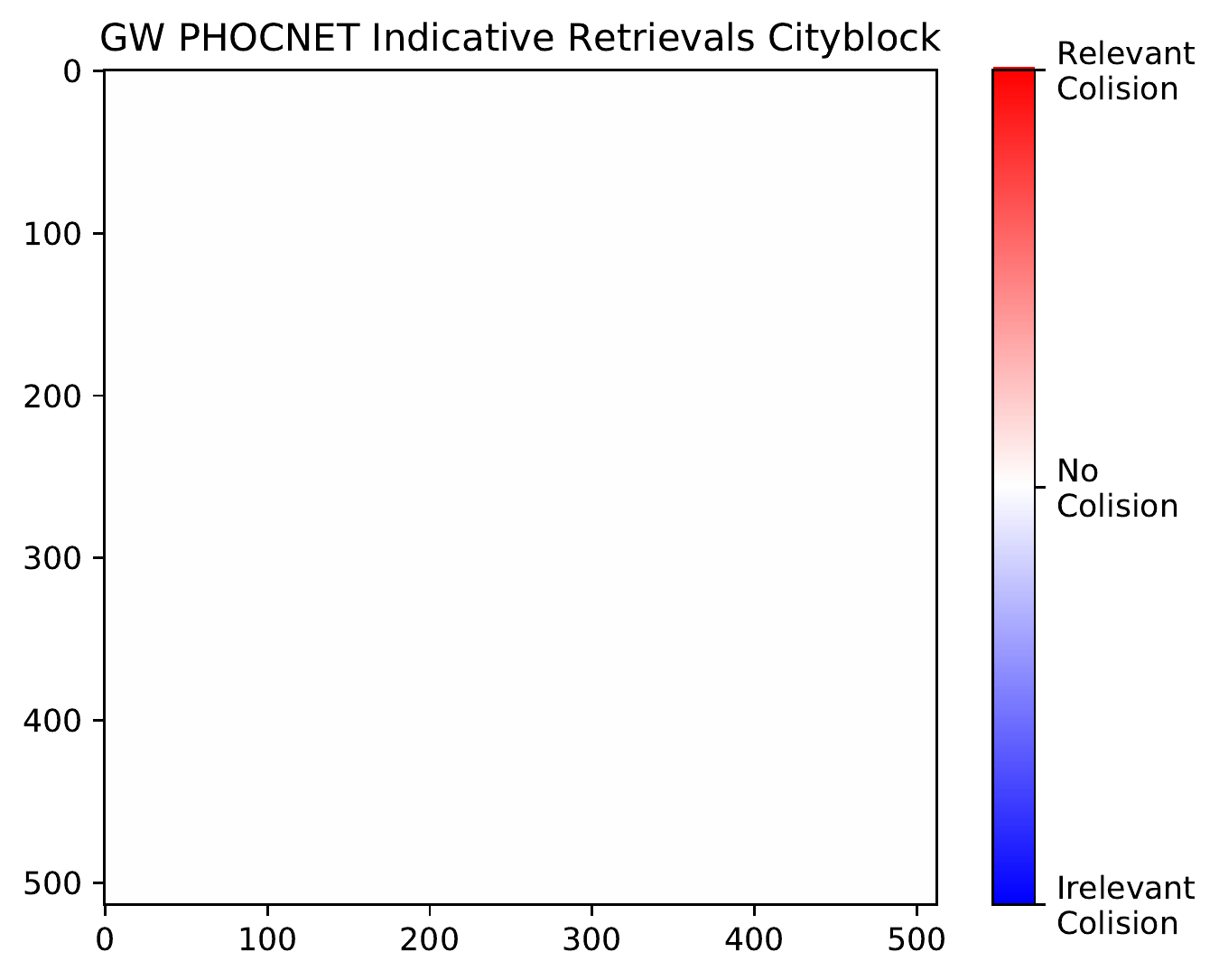}
\caption{PHOCNET embeddings for GW under various distances with $\epsilon = 10^{-10}$. Rows represent queries and columns represent samples sorted from left to right by similarity to each query.}
\label{fig:gw}
\end{figure}

The PHOCNET~\cite{sudholt2016phocnet} is a deep CNN, which is trained with a regression loss to map word-image inputs to a PHOC space ($\mathbb{R}^{504}$).
The PHOC space is a metric space under the cosine distance.
For evaluation, we use the George Washington (GW) dataset~\cite{fischer2012lexicon}. Specifically, we evaluate the test-set in a leave-one-out-image out cross-evaluation, \ie each sample of the test set is compared with the remainder of the test-set.
The test-set is stemmed for short (3 characters or less), and numerals so that there are \num{1164} word images left belonging to \num{431} classes.
Singleton samples, samples that occur only once and therefore can not be both a query and in the retrieval database, 
are removed from the query set, which is reduced to 899 samples, but are retained in retrieval database.
In \cref{fig:gw} the distribution of collisions under different distance metrics is shown.

A collision refers to two samples in the database having exactly the same distance. For visualization purposes, this was extended to all distances smaller than a threshold $\epsilon$ in \cref{fig:gw} and \cref{fig:rnd}.
Note, for a better visualization, the first \num{650} samples are dropped from the test-set.
When measuring the mAP of retrieval with GW, we observed values between \SI{95.34}{\percent} and \SI{95.36}{\percent} attributed to equidistant samples.

\subsection{Random Embeddings}
\begin{figure}[tb]
\centering
\includegraphics[width=.32\textwidth]{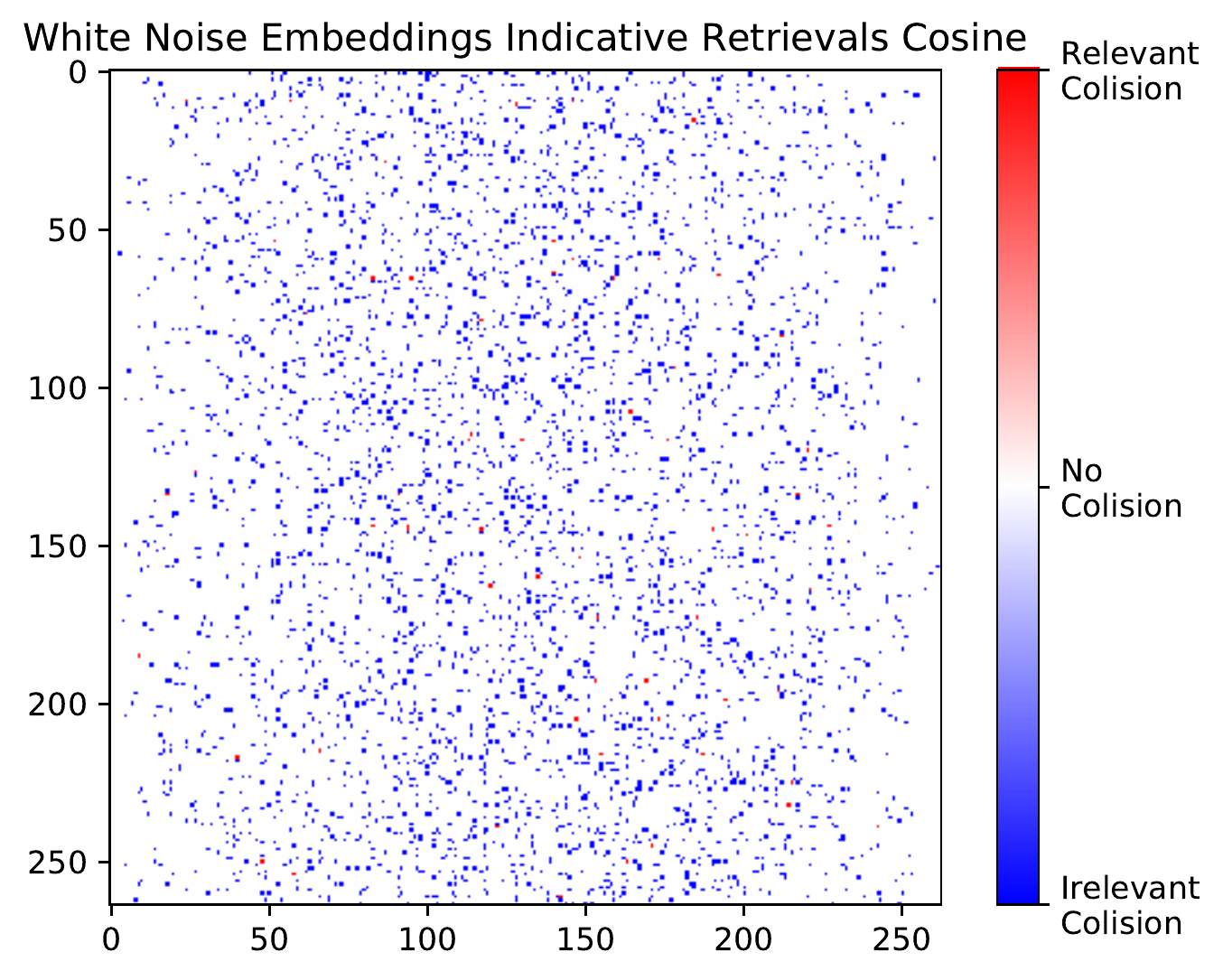}
\includegraphics[width=.32\textwidth]{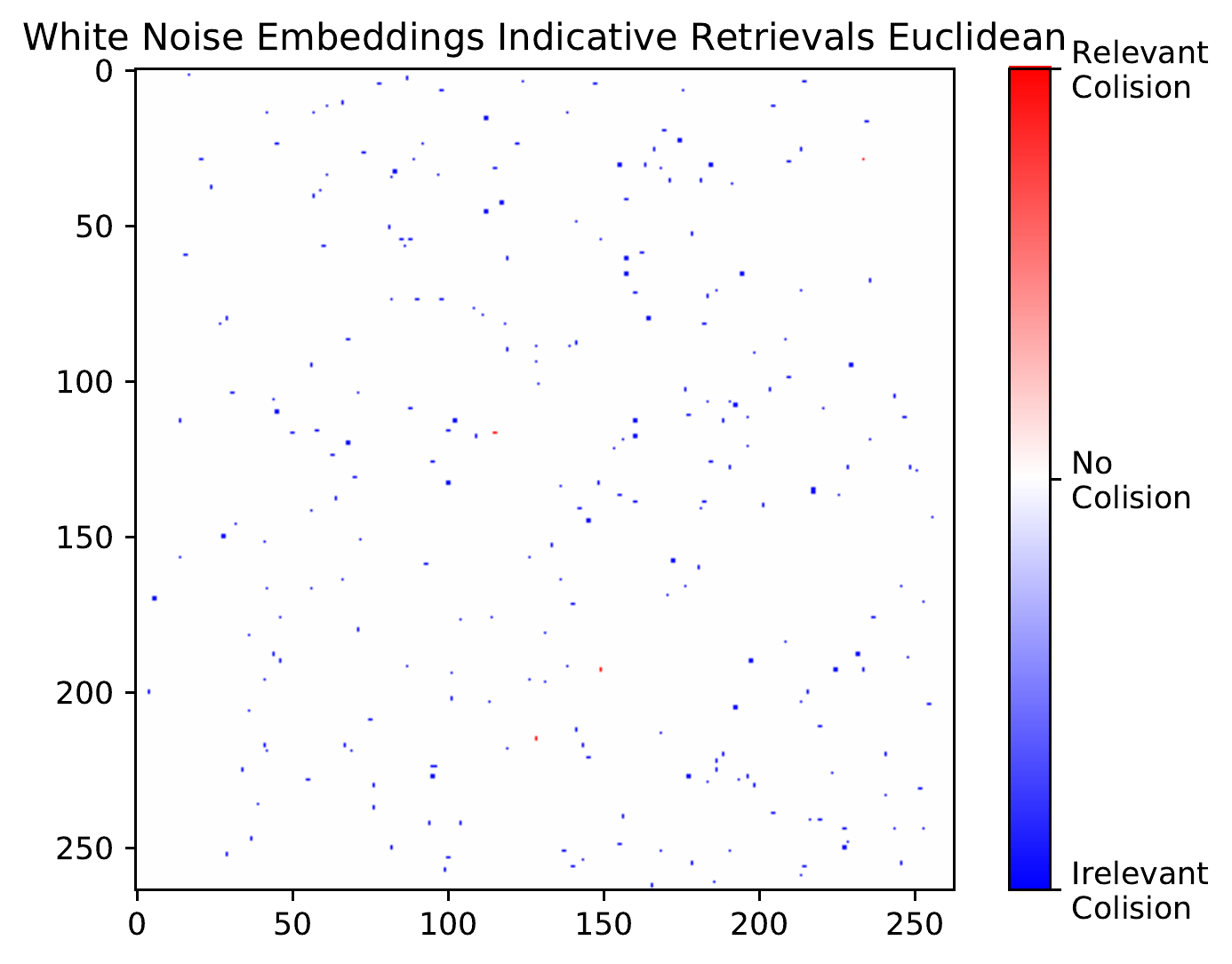}
\includegraphics[width=.32\textwidth]{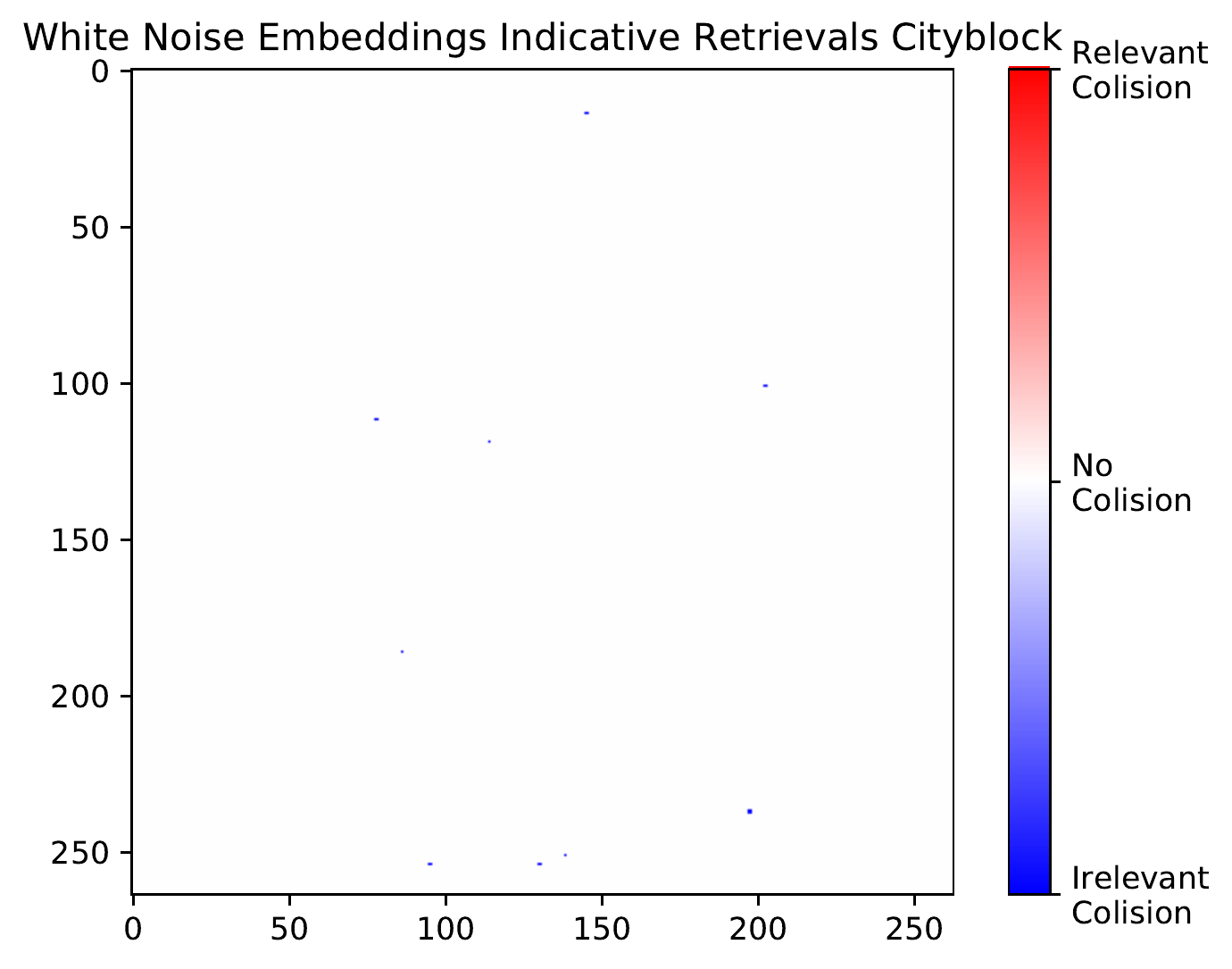}
\caption{Random embeddings with the same cardinalities and labels as GW.
Rows represent queries and columns represent samples sorted from left to right by similarity to each query.}
\label{fig:rnd}
\end{figure}

We perform an additional experiment to obtain further insights and contextualize the measurements on GW. 
We generated white noise embeddings of exactly the same cardinality as the PHOCNET embeddings uniformly sampled in the range $[0,1]$ having the same range as the PHOCNET embeddings.
We also used the same labels as GW test-set to make sure that the labeling statistics are identical.
In \cref{fig:rnd} we can see occurrence of collisions under different distance metrics.
In order to visualize collisions, we considered any consecutive samples having a difference greater than $\epsilon$ as colliding. In the case of random embeddings, in order to produce enough collisions to have plots comparable to GW  we had to increase $\epsilon$ to $10^{-6}$. 

\subsection{Analysis}
These experiments\footnote{All experiments and plots presented in this paper are reproducible and available at \url{https://github.com/anguelos/embedding_map}.} provide some insights into the described phenomena.
What stands out is the effect the different distance functions have on the same vectors.
Specifically it is worth observing that consistently across both GW and random embeddings, city-block distance produces a lot less collisions than Euclidean distance which in turn has approximately 20 times less collisions than cosine distance.
Conversely, it is worth pointing out that only in the case of the trained embeddings and cosine distance almost all collisions happen in the right side of the spectrum, where samples that are the furthest apart from each query concentrate.
Furthermore, PHOCNET in combination with cosine distance produces many collisions among relevant samples. This demonstrates the extent to which PHOCNET manages to regress perfect PHOC representations.
Another important observation is that in order to produce plots where collisions are visible when using random embeddings we had to increase the collision visibility criterion from $10^{-10}$ to $10^{-5}$.
The fact that a real-world method is \num{10000} times more prone to collisions than random data of the same cardinalities and distributions is a good indication that the probability of collision is practically unpredictable unless measured.  

\section{Proposed Solution}
\subsection{Determinism and Bounds}
The principal problem arising from equidistant embeddings is the unpredictability of their sorting.
The simplest remedy for this is to make the evaluation system consistently sort equidistant samples in the most favorable way possible.
Therefore, we define a new matrix $E$, which holds a small constant $\epsilon$ for any non-relevant element: 
\begin{equation}
E = (1-R)*\epsilon \;,
\end{equation}
where $R$ is the relevance matrix $R$, \cf \cref{metricestimation}.
Then, we can define two new similarity/distance matrices $D^+$ and $D^-$ as:
\begin{equation}
\begin{aligned}
D^+ = D + E \qquad
D^- = D - E\;.
\end{aligned}
\label{eq:optimist}
\end{equation}
In this way, the relevant and irrelevant matches are separated from each other. Note, collisions within the individual groups have no influence on the performance evaluation.

In order to only affect equidistant samples, $\epsilon$ must be smaller than the smallest observed difference between any pair of distances in any query that is greater than zero.\footnote{This can for example be easily achieved by switching from float to double precision and choosing $\epsilon$ appropriately.}

It follows that computing mAP from $D^+$ instead of $D$, provides an upper bound on all the plausible mAP estimates for given outputs of a system.
Respectively, by computing mAP from $D^-$, we can get the lower bound among all plausible mAP estimates of the performance of a system.
From here on we refer to the two bounds as mAP$^+$ and mAP$^-$.
\begin{figure}[tb]
\centering
\includegraphics[width=.47\textwidth]{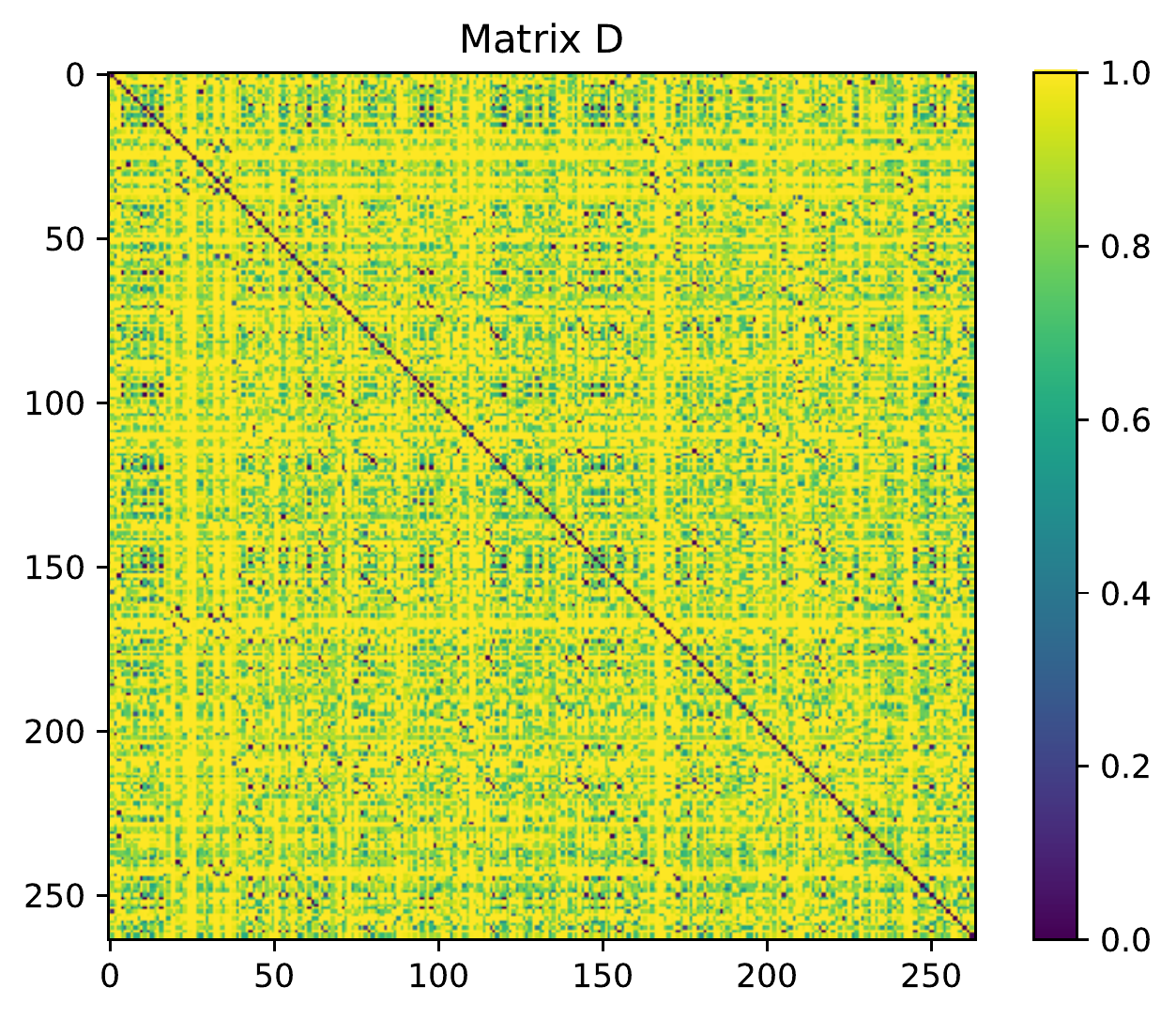}
\includegraphics[width=.47\textwidth]{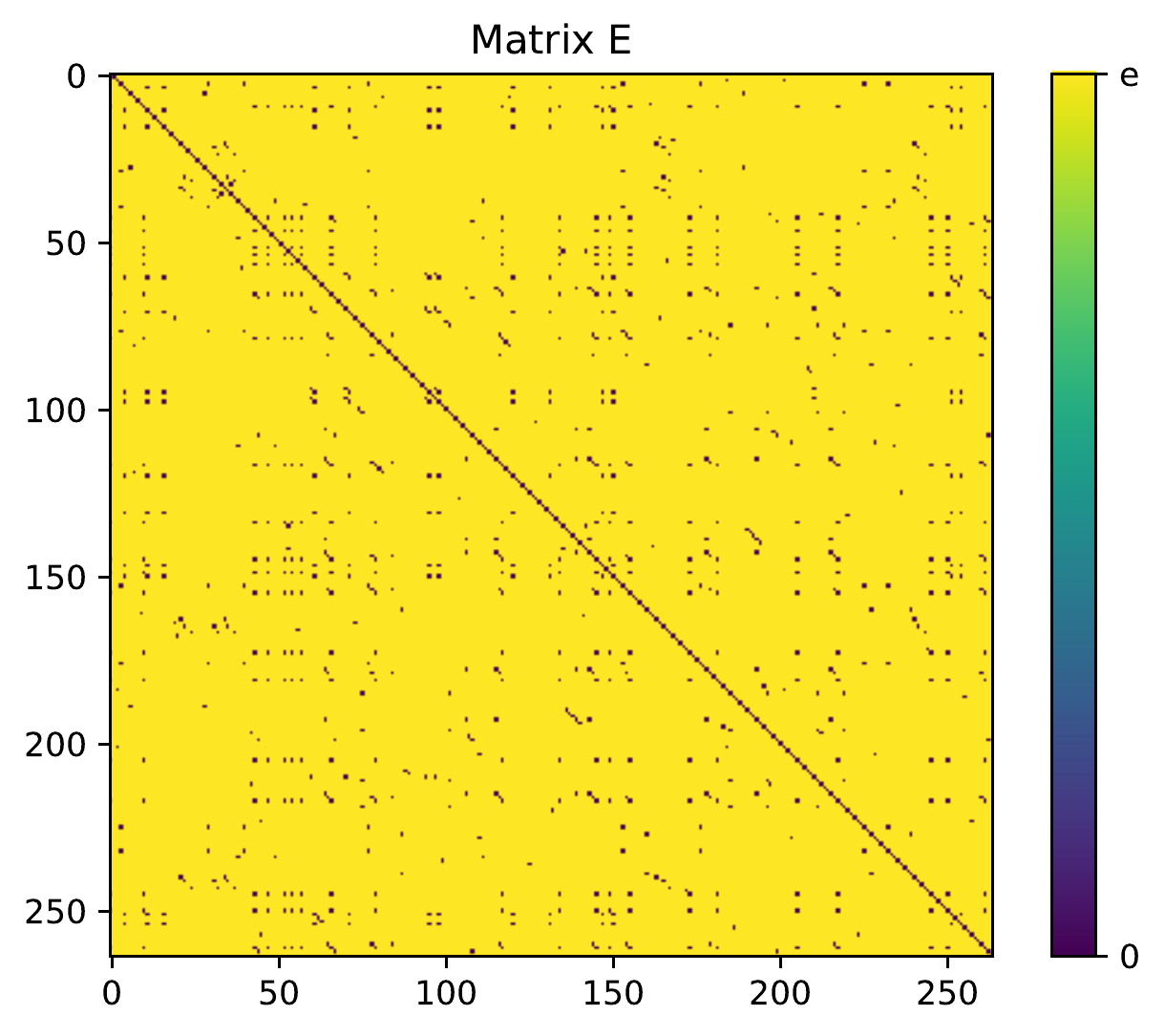}
\caption{Cosine distance matrix $D$ for the GW test-set and the matrix $E$.}
\label{fig:dm}
\end{figure}
\Cref{fig:dm} shows $D$ and $E$ computed on a part of the GW test-set.
It should be pointed out that the modality used is self-classification, where queries and database are the same samples, and the nearest sample to each query is always omitted as the sample is always itself with a distance of $0$.  
Given the small numerical effect equidistant samples have over the mAP in real world systems, providing the two bounds of mAP should be informative enough.
It should be pointed out that both bounds are deterministic with respect to the embeddings and the data labels, and therefore also their mean.
Nonetheless, the mean of the bounds is not directly related to the expected AP of a retrieval containing equidistant samples.

\subsection{Expectation}
While the bounds of all valid mAP are easy to compute, computing the expected mAP over all possible permutations of equidistant samples is not trivial.

As can be seen in \cref{fig:map_example} an ambiguous ranking contains one, or more sequences of equidistant samples that are both relevant and non-relevant.

By definition mAP only samples precision at the points where recall changes, thus equidistant sequences with only relevant or non-relevant sequences do not affect mAP and can be ignored.
It can also be deduced that the effect each sequence of equidistant samples has over the total mAP expectation can be independently computed for every equidistant sequence.

Each equidistant sequence of length $l$ containing $m$ relevant samples, is preceded by a retrieval of $n$ relevant samples from $k$ retrieved samples.
We then know that precision before the sequence is $n/k$ and after it is $(n+m)/(k+l)$.
The two fractions $(n+m)/(k+m)$ and $(n)/(k+l-m)$ are respectively the upper and lower bounds of all possible precision measurements occurring within the equidistant sequence.
By definition of the mAP, we also know that each equidistant sequence will affect the overall mAP $m$ times, thus $m$ is in effect a coefficient of the sequence.
Under the assumption of small equidistant sequences, the mean mAP for all possible permutations of equidistant samples could be computed but the brute force algorithm would be inefficient.

\section{Exploitation of the unpredictability}
\subsection{A computer security approach}
Although scientific work is predicated on the integrity of the scientists, it is important to keep in mind that there might be serious incentives for improving the perceived performance of a system.
The disqualification~\cite{bbc2016} of Baidu from the ImageNet competition~\cite{russakovsky2015imagenet} demonstrates that even the leading scientific teams can show ambiguous ethics.
More than that, the incident is an interesting example of how a participant to a competition can act in a manner that is ethically in a gray-zone rather than all-out cheating.
Performance evaluation design in the context of public competitions should have a computer security aspect to it, the rules and protocols should be designed in a way that ethical gray-zones are minimized.
People could always cheat or lie, but most people, will never cross that line.

\subsection{All zero embedding exploit}
Even though the experiments presented in \cref{experiments} demonstrate that the phenomenon of sorting ambiguity has a small effect under regular conditions, there are circumstances where it could be exploited and amplified the effect to an extreme level.

\begin{figure}[tb]
\centering
\includegraphics[width=.8\textwidth]{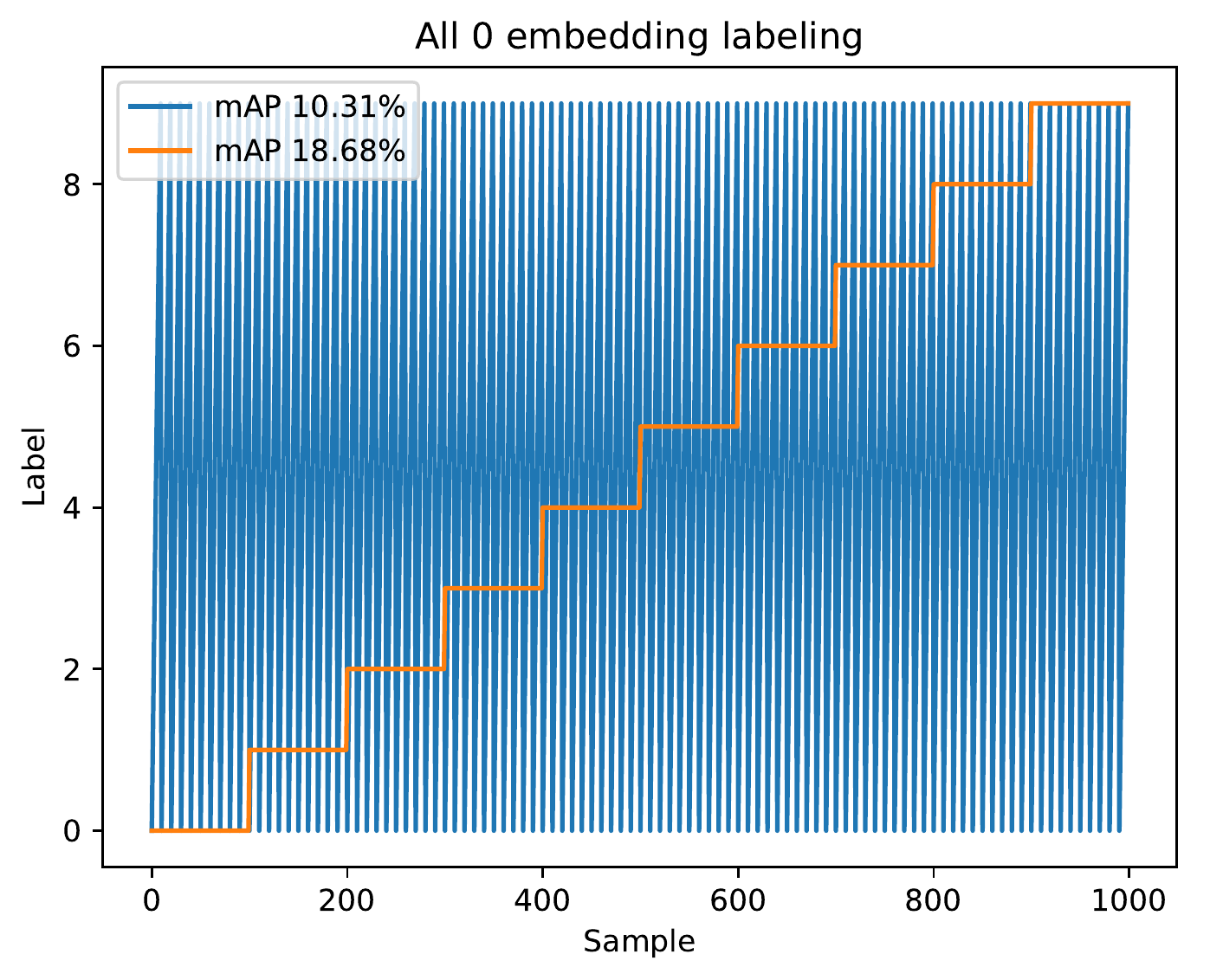}
\caption{Most favorable and least favorable orderings of the query samples.}
\label{fig:allzero}
\end{figure}

As a naive exploitation of the ambiguity we tried the following adversarial example.
We hypothesized a system that always maps any input sample to a vector of $\mathbb{R}^{1000}$ with all zeros.
When all embeddings have all zeros, then everything is equidistant.
All performance estimates depend on how the sorting algorithm deals with equal values.
We created labels for a thousand samples, \num{10} classes each having \num{100} samples.
Afterwards, we employed standard self-classification where query and retrieval samples are the same, also known as leave-one-out cross-validation.
By tweaking the order in which we evaluated the samples, which were identical, we managed to obtain different mAP measurements between \SI{10.32}{\percent} and \SI{18.68}{\percent}.
In~\cref{fig:allzero}, the sample arrangements that produce the most and least favorable mAP estimates are visible.
The exploit does not produce the full range of mAP$^-$ to mAP$^+$, which lies in the range \SI{5.18}{\percent} to \SI{100}{\percent}.
In order to produce such a variation, someone would probably need to alter the distance matrix $D$ instead of the order of the queries.
The exploit was not demonstrated on publicly deployed system but rather on a straight forward implementation of mAP as described in~\ref{metricestimation}.
The exploit is simpler to implement in leave-one-evaluation is also directly applicable to regular retrieval if the retrieval samples are sorted by class or if the third party being evaluated can infer how order of the samples relates to their classes. 

It should be pointed out that over \num{30} repetitions of random embeddings instead of all zero produced an mAP mean of \SI{10.4}{\percent} with a standard deviation of \SI{0.053}{\percent}.
The repercussions of this finding are quite significant as the adversarial all-zero system managed to outperform almost by two-fold the random system.
There are many cases where systems are considered state-of-the-art while marginally surpassing the random predictor.
For example, in gender identification from handwriting~\cite{djeddi2015icdar2015} the winner~\cite{nicolaou2015sparse} demonstrated a performance of \SI{62}{\percent} while the random predictor produces \SI{50}{\percent}.

\subsection{Protection from the exploit}
From the perspective of securing mAP against attacks exploiting equidistant samples, the simplest solution is to substitute mAP with mAP$^-$.
Adopting mAP$^-$ as the evaluation metric puts a penalty on equidistant samples without any foreseeable side-effect.
If someone being evaluated wants to be protected from map$^-$ having a penalty on him, he can easily avoid it by adding some noise on his outputs.
Given that equidistant samples occur rarely in regular conditions, having them in such abundance so that they affect the evaluation metrics significantly, should probably be attributed to intent or poor system design.
In either case this should not be rewarded.
The last but most important reason for adopting mAP$^-$ instead of mAP, is that no system that is agnostic to the inputs should ever out-perform a random predictor significantly.

\section{Conclusion}
\subsection{Key points}
The key points and arguments of this paper can be summarized as follows:
\begin{itemize}
\item{Randomness is allowed in evaluated systems but should not be accepted in evaluation metrics.}
\item{Unpredictability of an algorithm should not be treated as randomness.}
\item{Evaluation protocols should be treating systems as black boxes.}
\item{Embedding methods scale well large datasets and are the most applicable pattern recognition retrieval techniques.}
\item{Evaluation of embeddings requires sorting the distances from the query.}
\item{Relevant and non-relevant samples that are equidistant from the query make the exact mAP measurement unpredictable.}
\item{Although marginal, this phenomenon has been observed in real-world systems.}
\item{The unpredictability can be easily addressed but estimating the "True" mAP is more complicated.}
\item{The phenomenon can be used malevolently to demonstrate performance significantly better than the random predictor while totally independent from input data.}
\item{In the context of competitions and other rigorous testing, $mAP^-$ should be preferred over mAP as it penalizes the occurrence of ambiguities and motivates the method's creator to resolve them.}
\end{itemize} 

\subsection{Discussion}
The collision effect from equidistant samples in real scenarios is rather small, 
but it should be pointed out that detecting such phenomena in most 
circumstances is practically impossible. Therefore, we cannot really know how often they occur.
The fact that GPUs, which are used in practically all modern pattern recognition methods, operate on 32 bit floating point, makes equidistant sample ambiguity more plausible.
Moreover, embedding methods might produce near-discretized embeddings, such as the PHOC, or even discretized ones, such as POOF~\cite{berg2013poof}. This makes the occurrence of such phenomena even more probable than one would expect.
We believe that an evaluation metric should be robust against adversarial inputs and always provide meaningful results.
We also believe that the standard of reproducibility, to which an evaluation metric is set, should be higher than any other component of the experimental evaluation.
While one might argue that it is hard to prove the statistical significance of the analyzed phenomena, we believe that performance metrics should be held to the standard of algebra.
It can be argued that statistics are not as informative when analyzing phenomena such as numerical instability.
Numerical instability problems cannot be modeled as random variables because they are usually pseudo-deterministic; a system will usually be extremely consistent in producing the wrong number, thus repetition cannot provide an estimation on the distribution that the instability follows.

Computer and data science operates in a totally deterministic space where all randomness seizes at the point of digitization.
When the behavior of systems modeled black-boxes is the subject of scientific analysis, then the actual evaluation metrics are the principal mean of observation, it is important to know and control the exact amount of error that these metrics have.
Disparities on measurements of the same observation must be quantified, accounted for, and understood, as in many cases, one might be right to suspect they indicate a bug in his experimental pipeline.

By considering only statistically significant errors in the metrics as unacceptable, we let go of perfect reproducibility.

By providing an easily computable quantification of these effects, we remove this source of non-determinism from an otherwise purely deterministic experimental process.

\section*{Acknowledgments}
This work has been partially supported by the European fund for regional development, grant-nr.\ 211 and the Spanish project TIN2017-89779-P. The contents of this publication are the sole responsibility of the authors.
\bibliographystyle{splncs04}
\bibliography{bibliography}
\end{document}